\documentclass[a4paper, 10pt]{article}
\usepackage[a4paper, left=1in, right=1in, top=1in, bottom=1in]{geometry}
\usepackage{cite}
\usepackage{amsmath,amssymb,amsfonts}
\usepackage{algorithmic}
\usepackage{booktabs}
\usepackage{multirow}
\usepackage[table]{xcolor}
\usepackage{graphicx}
\usepackage{subcaption}
\usepackage{textcomp}
\usepackage{hyperref}
\usepackage[T1]{fontenc}
\usepackage{titlesec}
\usepackage{siunitx} 

\graphicspath{{assets/}}

\usepackage{fancyhdr}

\makeatletter
\newcommand{\toptitlebar}{\hrule height 3pt \vskip 1.3em}
\newcommand{\bottomtitlebar}{\vskip 1em \hrule height 1pt \vskip 2em}
\renewcommand{\@maketitle}{%
  \newpage
  \null
  \vskip -2.3em%
  \begin{center}%
    \toptitlebar
    {\Large\bfseries \@title \par}%
    \bottomtitlebar
    {\normalsize \@author \par}%
  \end{center}%
  \par
  \vskip 1.5em%
}
\makeatother

\fancypagestyle{firstpage}{
    \fancyhf{}
        \fancyfoot[L]{\parbox[t]{\textwidth}{%
        \hrule width 0.1\textwidth height 0.2pt \vspace{0.2em}
        \footnotesize{Code available at: \url{https://act-jepa.github.io}}}}
    
}

\titleformat{\paragraph}[runin]
  {\normalfont\normalsize\bfseries}
  {}{0pt}{}
\titlespacing*{\paragraph}{0pt}{1ex plus 0.5ex minus 0.2ex}{1em}

\title{
    ACT-JEPA: Novel Joint-Embedding Predictive Architecture for Efficient Policy Representation Learning
}

\author{
    \normalsize \textbf{{Aleksandar Vujinović \hspace{5em} Aleksandar Kovačević$^{*}$}} \\
    {\normalsize Faculty of Technical Sciences, University of Novi Sad} \\
    {\normalsize {\{aleksandar, kocha78\}@uns.ac.rs}} \\
    {\normalsize $^{*}$Corresponding author}
}

\begin{document}

\date{}
\maketitle
\thispagestyle{firstpage}

\begin{abstract}
Learning efficient representations for decision-making policies is a challenge in imitation learning (IL).
Current IL methods require expert demonstrations, which are expensive to collect.
Additionally, they are not explicitly trained to understand the environment.
Consequently, they have underdeveloped world models.
Self-supervised learning (SSL) offers an alternative, as it can learn a world model from diverse, unlabeled data.
However, most SSL methods are inefficient because they operate in raw input space.
In this work, we propose ACT-JEPA, a novel architecture that unifies IL and SSL to enhance policy representations.
It is trained end-to-end to jointly predict 1) action sequences and 2) latent observation sequences.
To learn in latent space, we utilize Joint-Embedding Predictive Architecture, which allows the model to filter out irrelevant details and learn a robust world model.
We evaluate ACT-JEPA in different environments and across multiple tasks.
Our results show that it outperforms the strongest baseline in all environments.
ACT-JEPA achieves up to 40\% improvement in world model understanding and up to 10\% higher task success rate.
Finally, we show that predicting latent observation sequences effectively generalizes to predicting action sequences.
This work demonstrates how integrating IL and SSL leads to efficient policy representation learning, an improved world model, and a higher task success rate.
\end{abstract}
\newcommand{\figarchitecturecaption}{%
\mathversion{normal}%
\textbf{ACT-JEPA overview.}
The model is trained end-to-end to predict action sequences and latent observation sequences, with both objectives optimized using $L_1$ loss.
It consists of four transformer-based components.
The context encoder $E_{\theta}$ encodes the current observation $o_t$ into a latent representation $s_x$.
The target encoder $E_{\overline{\theta}}$ encodes future observations $o_{t:t+n}$ into target latent representations $s_{y_{t:t+n}}$.
Then, the predictor $P_{\phi}$ takes $s_x$ and learnable mask tokens to predict target latent representations $\hat{s}_{y_{t:t+n}}$.
Finally, the action decoder $D_{\tau}$ takes $s_x$ and learnable mask tokens to generate a sequence of actions $\hat{a}_{t:t+n}$.
During inference, only $E_{\theta}$ and $D_{\tau}$ are utilized.
}

\begin{figure*}[ht]
    \centering
    \includegraphics[width=5in]{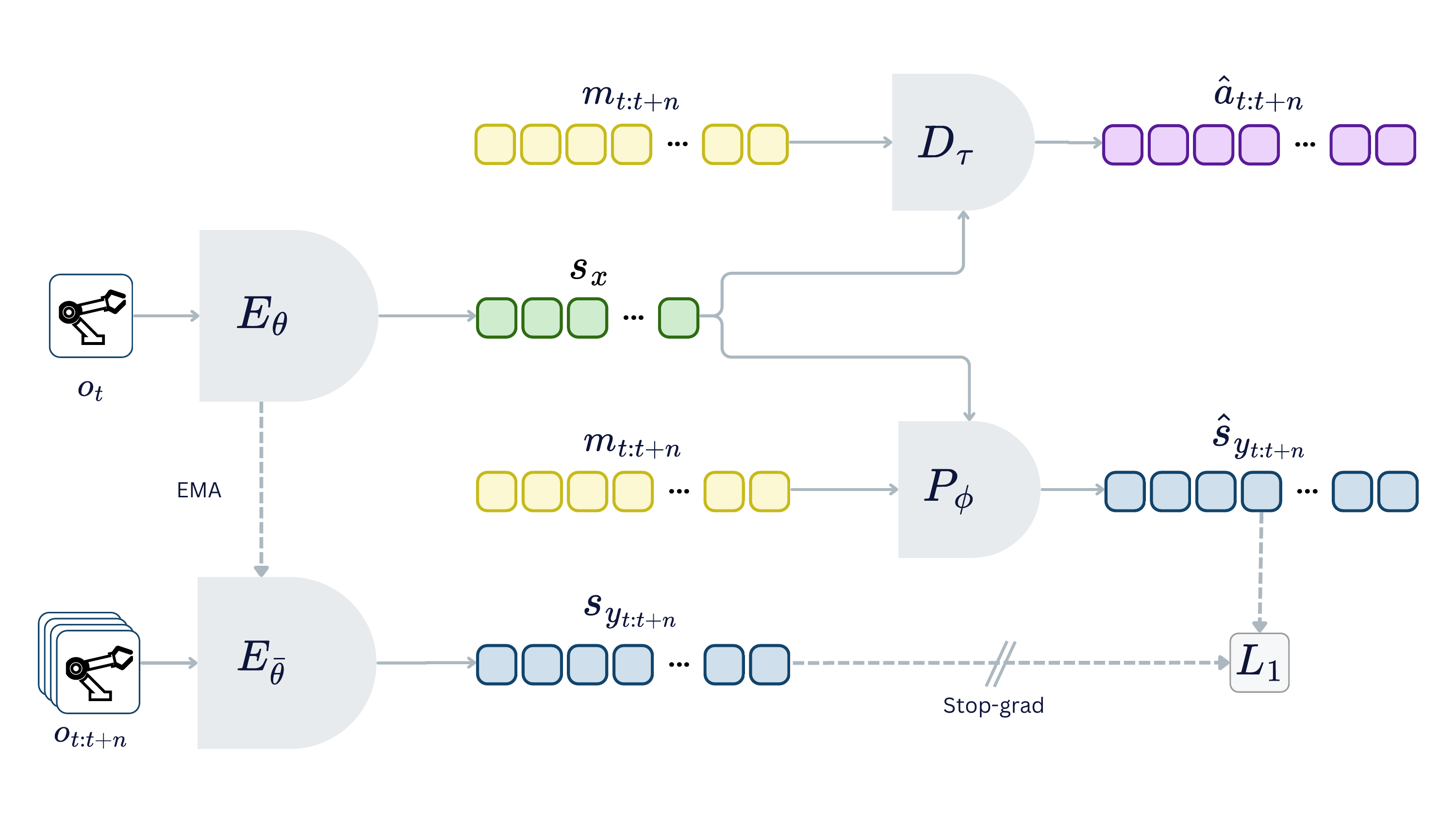}
    \caption{\figarchitecturecaption}
    \label{fig:overview}
\end{figure*}

\newpage
\section{Introduction}\label{introduction}

Learning end-to-end policies for decision-making tasks has long been a challenge in artificial intelligence.
Recent advances in imitation learning (IL) have shown strong performance, with models learning from expert demonstrations \cite{brohan_rt-1_2023,brohan_rt-2_2023,chi_diffusion_2024,zhao_learning_2023}.
While these models can learn to mimic expert actions, they do not explicitly learn a world model — how the environment evolves over time.
This limits their ability to predict future states and adapt to new situations \cite{garrido_learning_2024,ha_world_2018,hussein_imitation_2018,wu_masked_2023}.
Another key issue with current IL methods is their reliance on high-quality expert data, which is both costly and time-consuming to collect \cite{hussein_imitation_2018,vujinovic_using_2024,zhao_learning_2023}.
Learning only from expert demonstrations limits the model's ability to generalize across scenarios, including failure cases, reducing its ability to recover from mistakes or handle novel situations \cite{black_0_nodate,hussein_imitation_2018}.

Self-supervised learning (SSL) provides an alternative, as it does not require labeled data and allows a model to learn from a broader range of experiences, even failure cases.
However, most SSL methods make predictions in raw input space (e.g., pixel space), forcing the model to learn irrelevant or unpredictable details, which is computationally inefficient and requires large datasets \cite{lecun_path_nodate,assran_self-supervised_2023,bardes_revisiting_2024,lin_vedit_2024,yu_representation_2024}.
This makes SSL less practical for real-world applications, especially in complex domains such as robotics.

To address these limitations, we propose ACT-JEPA, a novel architecture designed to enhance policy representation learning.
ACT-JEPA unifies IL and SSL objectives to predict both action sequences and latent observation sequences. 
It is trained end-to-end, jointly optimizing both objectives.
The model learns to generate executable actions using IL, while simultaneously learning a world model using the SSL objective.
To learn a world model in latent space, we utilize Joint-Embedding Predictive Architecture (JEPA) \cite{lecun_path_nodate}. In contrast to SSL approaches that learn in raw input space, JEPAs learn by predicting in latent space, allowing the model to eliminate irrelevant and unpredictable information \cite{lecun_path_nodate,assran_self-supervised_2023,bardes_revisiting_2024,lin_vedit_2024,yu_representation_2024}.
This approach makes learning efficient, develops a robust world model, and improves representation quality.

IL methods, such as behavior cloning, face challenges like compounding errors due to their autoregressive nature \cite{brohan_rt-1_2023,brohan_rt-2_2023,zhao_learning_2023}. 
To resolve this, IL policies are often trained to predict action sequences instead of only one future action \cite{zhao_learning_2023,haldar_baku_2024}, which is also utilized in ACT-JEPA.
Building on this concept, we train ACT-JEPA to predict entire sequences of latent observations, instead of predicting only one future state \cite{cui_dynamo_2024,zhou_dino-wm_2024}.
Expanding the prediction horizon results in a world model with richer and more robust representations.
At the same time, learning in latent space makes the approach highly efficient.

We test our approach by running experiments across multiple tasks in simulated environments such as Push-T \cite{chi_diffusion_2024}, Meta-World \cite{yu_meta-world_2021}, and ManiSkill \cite{tao_maniskill3_2025}.
Our results show that ACT-JEPA outperforms other IL methods in all environments.
We first show that ACT-JEPA has up to a 40\% improvement in world model understanding compared to the strongest baseline.
Similarly, we test ACT-JEPA's performance in the environments and show that it has up to a 10\% higher task success rate.
Finally, we show that learning latent observation sequences successfully transfers to predicting action sequences.
These results show that ACT-JEPA is a promising direction for developing efficient, generalizable, and robust policy representations.

Learning robust policies with improved representations is particularly relevant in robotics, where sensor complexity and data volume are rapidly increasing.
Modern robotic systems (e.g., robodogs, humanoids, and dexterous hands) integrate high-dimensional inputs from joints, cameras, and tactile sensors \cite{bhirangi_anyskin_2024,lambeta_digitizing_nodate,zhang_whole-body_2024}.
These systems generate large amounts of data, but not all information is equally important; some is unpredictable or irrelevant for a given task.
By learning in latent space, ACT-JEPA can eliminate noise and focus on the most relevant features.
By utilizing the SSL objective, the model can reduce reliance on expert data and learn from diverse data, including failure cases.
Since the model is developed in latent space, it can naturally support various input modalities, making it well-suited for increasingly complex robotic platforms.

To summarize, our key contributions are:
\begin{itemize}
\item 
We present ACT-JEPA, a novel end-to-end architecture that learns to generate actions using imitation learning, while simultaneously learning a world model in latent space using JEPA. 
To the best of our knowledge, we are the first to combine IL and JEPA.
\item
By learning a latent world model, ACT-JEPA achieves up to 40\% relative improvement over IL baselines in world model understanding.
\item
Across multiple decision-making benchmarks, ACT-JEPA consistently outperforms established IL baselines, achieving up to 10\% absolute improvement in task success rate.
\item
Our joint optimization of IL and SSL world modeling outperforms both conventional two-stage training and IL-only policies under constrained data and computational budgets.
This is particularly important in real-world robotic deployments where policies are developed for specific tasks, rather than general-purpose settings.
\end{itemize}

The rest of the paper is organized as follows. In Section \ref{related-work}, we review related work. We present the ACT-JEPA architecture in Section \ref{act-jepa}. Section \ref{experiments} explains the experimental setup to test our architecture, presents results, and discusses our findings. Section \ref{conclusion} concludes our paper.
The training code and models are open source and available at \href{https://act-jepa.github.io}{\textnormal{\texttt{https://act-jepa.github.io}}}.

\section{Related work}\label{related-work}

\subsection{Imitation learning}
In imitation learning (IL), the agent learns from experts by imitating their behavior. 
The most common IL method is behavior cloning (BC), which formulates policy learning as a supervised learning task: the agent is trained to predict the expert's actions using a dataset of state-action pairs.
Many recent works have tried to improve policies by focusing on different architectures and objectives.
For example, some utilized large pretrained models and adapted them for downstream decision-making tasks \cite{brohan_rt-1_2023,brohan_rt-2_2023,kim_openvla_2024,shafiullah_behavior_2022}.
These models are autoregressive, trained to predict the next action token based on inputs like images and text instructions.
While they show promise, they are constrained by their autoregressive nature to predict the next token.
This leads to several limitations. First, prediction errors accumulate, which can push the model into states outside the training distribution \cite{zhao_learning_2023}. These models struggle with non-Markovian behavior, like pauses during demonstrations (e.g., when a human demonstrator stops to think or plan their next actions) \cite{zhao_learning_2023}. Continuous actions are discretized into predefined bins (tokens), requiring the selection of the optimal size and number of bins \cite{lee_behavior_2024,shafiullah_behavior_2022}. These models utilize a finite token vocabulary, which limits their ability to handle action spaces with a large or infinite number of actions. Finally, large models require remote hosting and cannot run on resource-constrained devices like drones or self-driving cars \cite{brohan_rt-2_2023}.

Other works have explored non-autoregressive generative models.
For example, some works utilized autoencoders \cite{zhao_learning_2023,haldar_baku_2024}, others used diffusion models \cite{chi_diffusion_2024,sridhar_nomad_2023,ghosh_octo_2024}, and recent works have explored flow matching models (closely related to diffusion models) \cite{black_0_nodate}.
These models are well-suited for continuous action data, eliminate the need for discretization, can handle non-Markovian problems, and help mitigate compounding errors.
The key is to predict an action sequence (i.e., action chunk) instead of a single action.
Despite the progress, these methods heavily rely on labeled expert action data for training, which is costly and time-consuming \cite{hussein_imitation_2018,zhao_learning_2023}.
Utilizing only expert data limits their ability to generalize to unseen scenarios or recover from failures, unless explicitly demonstrated during training \cite{black_0_nodate,hussein_imitation_2018}.

\paragraph{ACT.}
Action Chunking with Transformer (ACT) \cite{zhao_learning_2023} is a non-autoregressive transformer-based policy trained to predict action sequences from observations.
It can be implemented as a variational autoencoder (VAE) or a simpler autoencoder (AE).
By training the model to minimize the \(L_1\) loss on continuous action sequences, it avoids discretization and helps mitigate compounding errors and non-Markovian data. 
Among IL approaches, this one is the closest to our method, as both methods predict action chunks with a transformer-based architecture. However, ACT-JEPA additionally develops a world model and enables a deeper understanding of the environment from unlabeled data.
The specific ACT variant used in our experiments is described in \autoref{baselines}.

\subsection{Representation learning with Joint-Embedding Predictive Architectures}\label{representation-learning-with-joint-embedding-predictive-architectures}

Self-supervised learning (SSL) has revolutionized representation learning in domains such as natural language processing and computer vision. Through SSL, models can learn rich representations from a range of unlabeled data. Various objectives and transformer-based architectures have been explored, such as GPTs \cite{chen_generative_nodate,radford_language_nodate}, autoencoders \cite{he_masked_2021,tong_videomae_nodate}, and diffusion models \cite{ho_denoising_nodate,peebles_scalable_2023}. Their key drawback is a reconstructive objective: the model must predict complete representations, including irrelevant or unpredictable details \cite{assran_self-supervised_2023,cui_dynamo_2024,lin_vedit_2024,yu_representation_2024}. Consequently, these architectures are computationally expensive and require large training datasets, which results in lower quality learned representations. Joint Embedding Predictive Architectures (JEPAs) overcome these limitations by learning to predict in latent space, where the model is free to discard irrelevant and unpredictable details. This improves both representation learning and computational efficiency.

The first concrete JEPA implementation is the I-JEPA approach \cite{assran_self-supervised_2023}. The main idea is to mask parts of the input and predict the masked parts in latent space using a simple reconstruction loss in embedding space. For example, I-JEPA learns to predict masked blocks of images from the visible regions. Importantly, this is done without hand-crafted augmentations (e.g., scaling or rotation), which reduces inductive bias and provides a more general solution that can be applied beyond images. In video representation learning, the model learns to predict masked temporal tubes to capture spatial and temporal dynamics \cite{bardes_revisiting_2024}. In the audio domain, it has been used to extract meaningful features from spectrograms~\cite{fei_-jepa_2024}. In 3D data processing, it effectively learns representations from point-cloud data \cite{saito_point-jepa_2024}. It has also proven effective for learning touch representations in tasks such as slip detection and grasp stability~\cite{higuera_sparsh_nodate}. This broad applicability highlights JEPA's potential to drive improvements across a range of tasks and modalities.

\subsection{Self-supervised learning for policy representation}

In the context of policy learning, most SSL methods are focused on improving policy representation by learning good visual representations in pixel space \cite{bruce_genie_nodate,hu_gaia-1_2023,urain_deep_2024,yang_learning_2024}. Only recently have some SSL methods shifted focus towards learning environment dynamics in latent space.

\paragraph{DynaMo.} 
This approach employs a two-stage training process to first learn environment dynamics and subsequently learn a policy to output actions \cite{cui_dynamo_2024}. In the first stage, an encoder is pretrained on in-domain task images to learn environment dynamics (both inverse and forward dynamics models). To make this efficient, the environment dynamics are learned in latent space using the VICReg objective \cite{bardes_vicreg_2022}. In the second stage, the frozen pretrained encoder is leveraged to train an action decoder head with IL. Experiments showed that DynaMo's abstract representations are more robust than those learned by SSL methods operating in input space, outperforming the baselines in decision-making.

\paragraph{DINO-WM.} 
This method is trained to develop a world model from images in latent space, rather than pixel space \cite{zhou_dino-wm_2024}. It captures environment dynamics between two successive frames. Specifically, it predicts the latent representation of the next frame \(z_{t + 1}\), given the latent representation of the current frame \(z_{t}\), and the current action \(a_{t}\). The model is optimized in embedding space with \(L_2\) loss. To encode frames and obtain latent representations, DINO-WM uses pretrained DINO-v2, which achieves the best performance but is significantly larger than common backbones like ResNet-18. Instead of training an action decoder with IL, DINO-WM uses model predictive control (MPC), a common algorithm used in planning and control. Although this approach results in a robust world model, its pretraining requires access to ground-truth actions.

\paragraph{V-JEPA 2-AC.}
Most recently, V-JEPA 2-AC was released after the initial publication of ACT-JEPA \cite{assran_v-jepa2_2025}. It employs a two-stage approach to learn a general-purpose world model: first, JEPA is used to pretrain an encoder on general-purpose video datasets; then, the pretrained encoder is frozen, and a separate latent predictor model is trained on robotics data with ground-truth actions to learn environment dynamics.
Like DINO-WM, it uses MPC to sample actions.

While these studies show promise, they leave room for improvement. Their experiments primarily focus on comparing policy performance against SSL methods operating in pixel space. Moreover, their methods differ from ours in objectives, architectures, and overall goals. We summarize the main differences:
DynaMo adopts a two-stage process with separate pretraining and policy learning, whereas our method is trained end-to-end. This is generally more scalable and effective, as the learned representations are jointly optimized for downstream tasks; our experimental results also support this claim. Additionally, DynaMo relies on more complex pretraining procedures and objectives compared to our approach.
In contrast to DINO-WM and V-JEPA 2-AC, which rely on MPC for action selection, our approach directly learns actions through IL. Both MPC and IL have their respective strengths. However, IL can effectively operate in noisy real-world environments, whereas MPC performance is limited by the quality of a learned world model. Additionally, MPC is usually slower as it requires optimization at inference time. Finally, in contrast to these methods, ACT-JEPA predicts entire sequences of actions and latent observations, which mitigates compounding errors and develops richer representations.

While SSL has shown promise in representation learning across various domains, its application to policy learning remains underexplored. These challenges highlight the need for approaches that can efficiently improve internal representations, learn a world model, and enhance decision-making. ACT-JEPA addresses these gaps by combining IL and SSL to learn action sequences and latent observation sequences end-to-end.
\section{ACT-JEPA}\label{act-jepa}
In this section, we describe ACT-JEPA, illustrated in ~\autoref{fig:overview}. 
It unifies IL and SSL objectives to learn executable actions and develop a world model.
We train it end-to-end, jointly optimizing both objectives.
To learn executable actions, the model is trained to predict a sequence of \(n\) future actions using IL.
Predicting action sequences, instead of a single future action, improves performance and mitigates issues such as compounding errors and non-Markovian behavior in the data \cite{zhao_learning_2023}.
To develop a world model, the model is trained to predict latent observation sequences using SSL.
Predicting entire sequences, instead of a single future state, improves world model understanding.
At the same time, learning in latent space makes this approach highly efficient; to achieve this, we build on the recent JEPA architecture.
Notably, SSL is used to learn how the environment evolves, not to imitate actions; for example, even if a non-expert demonstration shows a car driving off a cliff, the model does not learn to perform those actions. Instead, it learns the states leading to that scenario, improving its world model.
In the following, we describe the theoretical background, the main components of our architecture, and the objectives.

\subsection{Background}\label{background}

In this section, we provide the theoretical foundation for the ACT-JEPA architecture. We introduce key concepts in robot learning, such as imitation learning and behavior cloning. Then, we describe JEPA as a novel self-supervised approach.

\subsubsection{Policy learning}\label{policy-learning}

Imitation learning methods train a policy \(\pi\) on a dataset of expert demonstrations \(\mathcal{D}\) to mimic the expert's behavior. Each demonstration (episode) represents the full interaction from the start step \(t = 0\) to the end (termination) step \(t = T\). A demonstration consists of actions and observations. All episodes in the dataset are successful, meaning that the given task was successfully solved by the expert.
Behavior cloning is a common algorithm that casts imitation learning as supervised learning, mapping observations to actions. The policy \(\pi\) produces action(s) given some observation(s) to minimize the difference between the predicted actions and the expert's actions. In general, the policy can be defined as \(\pi\left( a_{t:t + n}\mid o_{t} \right)\), producing a sequence of \(n\) future actions. At test time (inference), the policy operates iteratively: every \(n\) steps, it receives observation(s) \(o_{t}\), generates a sequence of \(n\) future actions \(a_{t:t + n}\), and executes them.

\subsubsection{Joint-Embedding Predictive Architecture}\label{joint-embedding-predictive-architecture}

Self-supervised learning (SSL) is a representation learning method in which the system learns useful representations from its inputs. The Joint-Embedding Predictive Architecture (JEPA) is a novel approach within SSL that learns to produce similar embeddings for compatible inputs \(x\) and \(y\), and dissimilar embeddings for incompatible inputs \cite{assran_self-supervised_2023}. The loss function is applied between embeddings in latent (abstract) representation space, not raw input space (e.g., pixel space). Consequently, this elevates the level of abstraction and brings benefits such as improved training time and generalization.

In JEPAs, the model architecture consists of three main components: context encoder, target encoder, and predictor, as in \autoref{fig:overview}. The context encoder \(f_{\theta}\) takes an input \(x\) and produces a context representation \(s_{x}\). The target encoder \(f_{\overline{\theta}}\) encodes a target \(y\) and produces a target representation \(s_{y}\). To predict the target representation \(s_{y}\), the predictor \(g_{\phi}\) takes in the output of the context encoder \(s_{x}\) and a (possibly latent) variable \(z\), and outputs the predicted target representation \({\hat{s}}_{y}\).
For example, in the I-JEPA implementation \cite{assran_self-supervised_2023}, the context encoder was utilized to process one segment of an image, while the target encoder processed another segment (e.g., a missing segment of an image). The predictor then used the context representation to predict the representation of the missing segment.
The loss function is usually implemented as the \(L_{1}\) or \(L_{2}\) distance between the target representation \(s_{y}\) and the predicted target representation \({\hat{s}}_{y}\) \cite{assran_self-supervised_2023,bardes_revisiting_2024}. To train a JEPA model, the parameters of the context encoder \(\theta\) and the predictor \(\phi\) are typically updated with gradient descent. The target encoder is often implemented as a copy of the context encoder and is updated via exponential moving average (EMA) of the context-encoder parameters.

\subsection{Architecture overview}\label{architecture-overview}
ACT-JEPA is an end-to-end architecture that efficiently learns representations relevant for decision-making and world model understanding (\autoref{fig:overview}). The architecture consists of four main components: context encoder, target encoder, predictor, and decoder. At the core of each component is a transformer architecture \cite{vaswani_attention_2017}. All components are utilized during training. During inference, only the context encoder and decoder are utilized to generate action sequences, discarding the target encoder and predictor.

The inputs to the model are observations of different modalities. In our implementation, we set the context encoder to receive a current image and the proprioceptive state, while the target encoder receives a sequence of proprioceptive states. We note that the architecture is more general and can support various modalities as inputs (e.g., images, depth images, proprioceptive states, pose markers, tactile feedback, images from different camera views).

\subsubsection{Context encoder}\label{context-encoder}

The context encoder \(E_{\theta}\) takes as input observations at a timestep \(t\) and outputs a latent representation of the environment \(s_{x}\). The output \(s_{x}\) is shared between the tasks of predicting action sequences and latent observation sequences, which requires the context encoder to capture information relevant to both tasks. This design improves the model's internal representations, understanding of environment dynamics, and decision-making.

We implement the context encoder to receive an image, the proprioceptive state, and a task label. Each modality is encoded with a dedicated function to obtain modality-specific tokens. For instance, proprioceptive states are encoded using linear layers. Task labels are one-hot encoded. To process images, we follow the common approach of utilizing the pretrained ResNet-18 model to extract feature maps \cite{cui_dynamo_2024,haldar_baku_2024,lee_behavior_2024,zhao_learning_2023}. The feature maps are flattened along the spatial dimension to form a sequence of tokens, with positional encoding applied to preserve spatial relationships. ResNet-18 is chosen for its simplicity and effectiveness. While our approach aligns with standard architectures, the flexibility of our architecture allows for the integration of more advanced backbones such as DINO-v2 \cite{oquab_dinov2_2024,zhou_dino-wm_2024} or conditioning mechanisms like FiLM \cite{brohan_rt-2_2023,haldar_baku_2024,perez_film_2017}, which may yield improved performance in more complex tasks. Once all modalities are encoded into tokens, they are concatenated into a single sequence and fed into the transformer model. The model outputs the latent representation \(s_{x}\), which serves as the input to the subsequent decoder and predictor components.

\subsubsection{Target encoder}\label{target-encoder}

The target encoder receives an observation sequence \(o_{t:t + n}\) as input and outputs a sequence of latent observations \(s_{y_{t:t + n}}\). These representations capture how the environment evolves over time. They also serve as the targets for the second objective (predicting latent observation sequences). By representing targets in latent space, the model eliminates irrelevant or unpredictable details from the target representation.

We encode the input sequence with a modality-specific projection function. We select proprioceptive states as the target modality and encode them with a linear projection layer (as was done in the context encoder). Then, positional encoding is added to preserve temporal information. The encoded tokens are fed into the transformer model to process them. The output is a sequence of tokens \(s_{y_{t}},\ldots,s_{y_{t + n}}\), where each token is a latent observation. Thus, the whole output \(s_{y}\) represents the sequence of latent observations.

By outputting a sequence, the target encoder captures temporal environment dynamics. In contrast, the context encoder captures a representation of the environment at the current timestep \(t\) only. This enables the target encoder to produce non-trivial and semantically meaningful targets, while the context encoder provides informative context for downstream tasks. Under these settings, we align with prior works to design non-trivial prediction tasks while maintaining an informative context.

\subsubsection{Predictor}\label{predictor}

The predictor \(P_{\phi}\) learns how the environment evolves over time. It receives two inputs: the output of the context encoder \(s_{x}\) and a sequence of \(n\) mask tokens \(m_{t:t + n}\). Each mask token is a learnable vector and corresponds to an abstract observation we wish to predict. Instead of simply concatenating both inputs and passing them through a transformer model, we use a cross-attention block to condition on the context \(s_{x}\). This enriches the mask tokens with contextual information before they are processed by a transformer model. This is in contrast to similar JEPA architectures that use self-attention over all inputs \cite{assran_self-supervised_2023,bardes_revisiting_2024,cui_dynamo_2024}. The cross-attention blocks have linear complexity with respect to context length, allowing our model to efficiently handle longer sequences \cite{bar_navigation_2024}. By using cross-attention, we achieve a more flexible and efficient processing mechanism, as we reduce the number of tokens processed by the subsequent transformer blocks.
The output of the predictor is a sequence of tokens \({\hat{s}}_{y_{t}},\ldots,{\hat{s}}_{y_{t + n}}\), where each token represents the predicted latent observation. We select proprioceptive states as the observation modality, consistent with the modality used in the target encoder.

\subsubsection{Decoder}\label{decoder}

We utilize the action decoder \(D_{\tau}\) to predict action sequences. It takes in the output of the context encoder \(s_{x}\) and a sequence of \(n\) mask tokens \(m_{t:t + n}\). Each mask token is a learnable vector and corresponds to an action we wish to predict. 
As in the predictor, we use a cross-attention block to condition on \(s_{x}\), adding contextual information to the mask tokens, which are then processed by a transformer model. The output is a sequence of predicted actions to be executed in the environment \({\hat{a}}_{t},\ldots,{\hat{a}}_{t + n}\). 

The independent action decoder adds flexibility to the architecture. This is aligned with similar JEPA works that use independent decoder heads for downstream tasks (e.g., classifying images or generating pixels) \cite{assran_self-supervised_2023,bardes_revisiting_2024}. In our context of generating actions, it allows for easy substitution with more sophisticated generative models, such as diffusion \cite{chi_diffusion_2024,haldar_baku_2024}. However, more complex action decoders are often unnecessary if the encoder outputs good representations \cite{haldar_baku_2024}.

\subsection{Objective}\label{objective}

We train the model with two objectives: given a current observation \(o_{t}\), predict 1) an action sequence \({\hat{a}}_{t},\ldots,{\hat{a}}_{t + n}\) and 2) a latent observation sequence \({\hat{s}}_{y_{t}},\ldots,{\hat{s}}_{y_{t + n}}\). The first objective is supervised, requiring expert actions, while the second objective is self-supervised, allowing the model to learn from unlabeled observation data. These objectives ensure the model learns both low-level actions and high-level environment dynamics. To optimize both objectives, we utilize two loss functions.

\paragraph{Action loss.} For the first objective, we aim to reconstruct a sequence of future actions. Predictions are made in the given action space. To achieve this, we utilize \(L_1\) loss, following \cite{zhao_learning_2023}. The loss penalizes distances between the predicted and the target actions: 
\begin{equation}
    \mathcal{L}_{actions} = \frac{1}{n} \sum_{i=0}^{n} \left\| \hat{a}_{t+i} - a_{t+i} \right\|_1.
    \label{eq:action_loss}
\end{equation}

\paragraph{Observation loss.} For the second objective, \(L_1\) is also utilized as it was the most efficient in our experiments. Here, predictions are made in latent space, rather than the given observation space. Therefore, the loss penalizes the distance between the predicted latent observations and their corresponding targets: 

\begin{equation}
    \mathcal{L}_{observations} = \frac{1}{n} \sum_{i=0}^{n} \left\| \hat{s}_{y_{t+i}} - s_{y_{t+i}} \right\|_1.
    \label{eq:observation_loss}
\end{equation}

\paragraph{Final loss.}
The model is trained end-to-end by jointly optimizing both objectives. The final loss is computed as the sum of the two individual losses: 
\begin{equation}
    \mathcal{L} = \mathcal{L}_{actions} + \mathcal{L}_{observations}.
    \label{eq:final loss}
\end{equation}

This is in contrast to most approaches that separate SL and SSL into two distinct training stages \cite{assran_self-supervised_2023,bardes_revisiting_2024,higuera_sparsh_nodate,black_0_nodate}. In the two-stage approach, the model is first pretrained with SSL to learn useful representations, and is then adapted to a downstream task using SL, often through a task-specific decoder (e.g., image classification). Most of the model's knowledge is acquired during the pretraining stage, but this requires large amounts of pretraining data and computational resources. In our experiments, we found that combining both objectives into a single training stage is crucial for achieving strong performance. Jointly optimizing both losses in an end-to-end manner proved more effective than the two-stage approach.

Following the prior JEPA works \cite{assran_self-supervised_2023,bardes_revisiting_2024,fei_-jepa_2024}, the parameters of the context encoder, predictor, and decoder \((\theta,\ \phi,\ \tau)\) are learned through gradient-based optimization. The parameters of the target encoder \(\overline{\theta}\) are updated with the exponential moving average of the context encoder parameters, which is necessary to prevent representation collapse in latent space.
\section{Experiments}\label{experiments}

\newcommand{\RQone}{Does ACT-JEPA improve world model understanding by learning latent observation sequences?}
\newcommand{\RQtwo}{Does ACT-JEPA improve decision-making compared to established baselines?}
\newcommand{\RQthree}{Does learning a world model with JEPA capture features useful for policy learning?}
\newcommand{\RQfour}{How does joint optimization of IL and SSL objectives compare to optimizing them separately?}

In this section, we evaluate ACT-JEPA's performance in world model understanding and decision-making. We first describe environments used for evaluation. Next, we outline the baselines used for comparison. Finally, we present experiments designed to answer the following research questions:
\begin{itemize}
	\item \RQone
	\item \RQthree
	\item \RQfour
	\item \RQtwo
\end{itemize}

\subsection{Environments}\label{environments}

\begin{figure*}[t]
    \centering
    \includegraphics[width=0.99\textwidth]{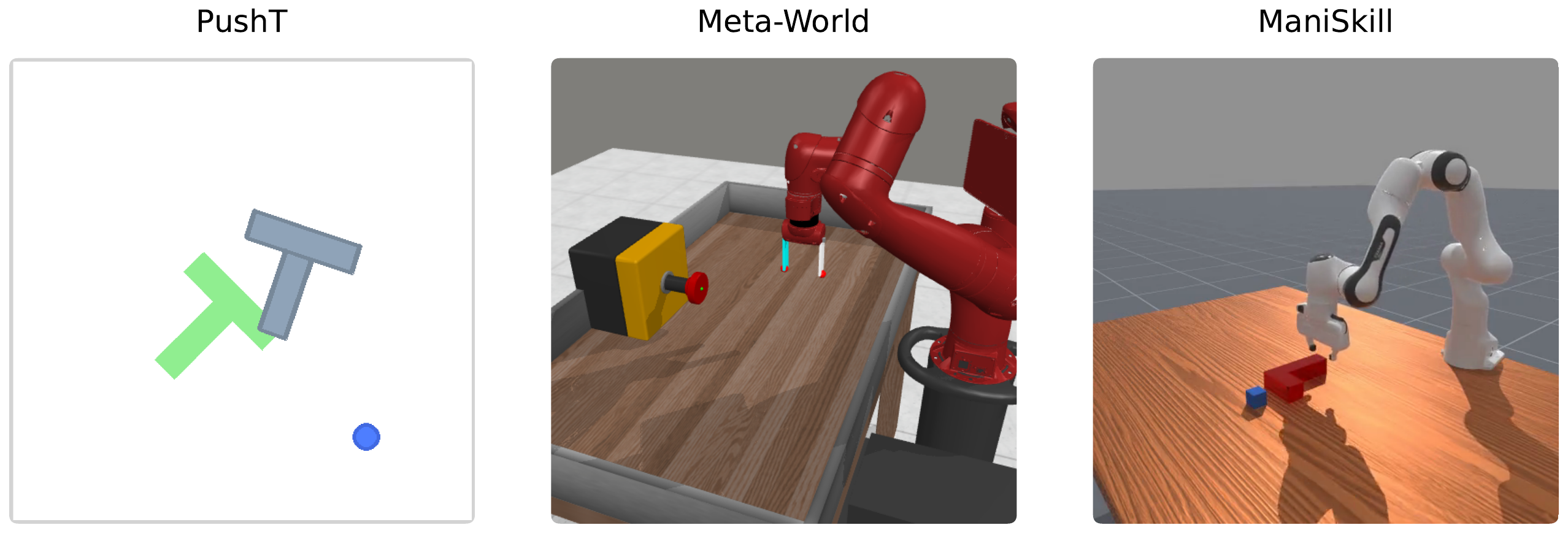}
    \caption{
    \textbf{Benchmark environments used for evaluation.} ACT-JEPA is evaluated on Push-T, Meta-World, and ManiSkill, covering a range of 2D and 3D robotic manipulation tasks.
    }
    \label{fig:benchmark_envs}
\end{figure*}

We describe the environments in which we test the hypotheses. All tasks in the environments have continuous observation and action spaces.
\autoref{fig:benchmark_envs} shows representative observations from the benchmarks.

\paragraph{Push-T.}
This is a 2D environment in which an agent (circle) has to push a moveable T-shaped object to a fixed T-shaped target area \cite{chi_diffusion_2024}.
The action space and proprioceptive space are two-dimensional.
This task requires precise object manipulation as the task is considered solved when the T-shaped object covers the target area.
The dataset consists of 206 human-collected demonstrations with RGB images of size (96, 96). 
Due to the low resolution, which makes fine-grained coverage difficult to assess reliably, we define the task as successful when the object covers at least 90\% of the target area.

\paragraph{Meta-World.}
This environment suite consists of a diverse set of tasks that share the same agent and workspace \cite{yu_meta-world_2021}.
It combines motions (e.g., reach, push, grasp) with varying objects to create tasks such as opening a drawer, opening a door, or pushing a button.
This setup ensures a shared underlying connection between the tasks, as they share similar environment dynamics.
This connection is crucial as it allows for representations learned in one task (e.g., opening a drawer) to transfer effectively to another (e.g., opening a door).
The robot itself has four joints, therefore the proprioceptive state and the action space are four-dimensional.
We used RGB images of size (128, 128) to balance computational efficiency with sufficient resolution for extracting meaningful features, as this proved to be effective in our experiments.
We selected 15 different tasks and utilized a scripted policy to collect the training data.
To focus on generalization and shared environment dynamics, we selected tasks that rely on a common set of manipulation primitives (e.g., reaching, pushing, opening), while excluding tasks that do not share these core operations (e.g., kicking a ball into the goal).
We collected 40 demonstrations per task, each with a different initial state (e.g., varying object positions).

\paragraph{ManiSkill.}
ManiSkill is a suite of simulated 3D robot manipulation tasks designed to evaluate object manipulation and generalization in diverse scenarios \cite{tao_maniskill3_2025}.
To ensure a consistent evaluation framework, we filtered the tasks by first identifying tasks with available expert demonstration datasets.
We then narrowed this selection to those utilizing the same robot (a two-finger gripper arm) and the same action space.
This resulted in five tasks (pick cube, push cube, stack cube, pull cube with a tool, and insert a peg sideways).
We used 50 demonstrations per task with RGB images of size (128, 128).

\subsection{Baselines}\label{baselines}

We compare ACT-JEPA with two behavior cloning methods, representing well-established baselines. To provide a fair comparison with the proposed architecture, we chose baseline policies that have the following in common: (1) visual data is preprocessed by a CNN-based backbone, (2) the transformer architecture is at the core of each component, and (3) each policy model was designed to directly output actions in continuous space. In the following, we describe the baselines in detail.

\paragraph{Autoregressive (AR) transformer.}
This is a GPT-style autoregressive (AR) transformer decoder policy.
The input is a sequence of previous actions and observations (e.g., images) represented as tokens.
The policy predicts the next action based on this sequence, leveraging the transformer's attention mechanism to effectively model long-term dependencies.
It is trained to minimize the \(L_2\) loss between the predicted and actual action.
This baseline is equivalent to Decision Transformer (DT) \cite{chen_decision_nodate}, but without using rewards as input.
It is also similar to Behavior Transformer (BeT) \cite{shafiullah_behavior_2022}, but without action discretization. 
Both DT and BeT are trained from scratch and have been widely adopted in similar settings.
Thus, we choose the AR transformer baseline, as it requires minimal modifications to the original GPT framework.
This makes it a relevant and effective baseline.

\paragraph{ACT.}
Another baseline we utilize is Action Chunking with Transformer (ACT) \cite{zhao_learning_2023}.
We implement ACT and ACT-JEPA to match in terms of inputs, outputs, and the training objective, enabling a controlled comparison.
We use the AE variant instead of VAE, removing the variational encoder component to simplify the architecture without sacrificing performance in our experiments.
This architecture is similar to the recent BAKU architecture \cite{haldar_baku_2024}, which achieves state-of-the-art results in similar environments.
However, we do not add BAKU's additional extensions, such as observation history, FiLM conditioning \cite{perez_film_2017}, or complex text-based conditioning.
While it is possible to implement these modifications, we intentionally omit them to maintain the focus on learning a world model, not on incorporating additional features that may distract from this objective.
Therefore, ACT is the most relevant baseline for our method. 
The key architectural difference is that ACT uses only IL, while ACT-JEPA additionally utilizes JEPA to learn a world model.

\subsection{\RQone}
Effective policy representations should capture information relevant for action prediction and world modeling.
ACT-JEPA is designed to address this by learning both action sequences and latent observation sequences.
To assess whether learning latent observation sequences improves world model understanding, we test if the learned representations can be used to predict future states.
We evaluate representation quality using a probing task \cite{bardes_revisiting_2024}.
For each trained policy, we freeze the context encoder, discard other components, and train a randomly initialized decoder to reconstruct future states from the frozen representations.
The observation modality is proprioceptive states, as this was the selected target modality in our experiments.
To evaluate performance, we use Root Mean Squared Error (RMSE) and Absolute Trajectory Error (ATE) as metrics \cite{sturm_evaluating_nodate,bar_navigation_2024}. 
RMSE highlights larger deviations by penalizing them more, giving an overall measure of error magnitude, while ATE measures how far off the predicted trajectory is from the ground truth trajectory. 
Together, these metrics offer a comprehensive evaluation of both the magnitude and alignment of the trajectory.

Our findings are presented in \autoref{tab:world_model_eval}.
ACT-JEPA consistently outperforms the ACT baseline across all benchmarks, reducing the prediction error by 29--37\% (RMSE) and 29--40\% (ATE).
These consistent improvements across diverse tasks demonstrate that ACT-JEPA learns more efficient and accurate world model representations.

To understand how representations evolve during training, we repeat the experiment at regular intervals throughout training.
This approach reveals not only the final performance, but also how representation quality improves over time.

\begin{figure*}[t]
    \centering
    \begin{minipage}[b]{0.48\textwidth}
        \centering
        \includegraphics[width=\textwidth]{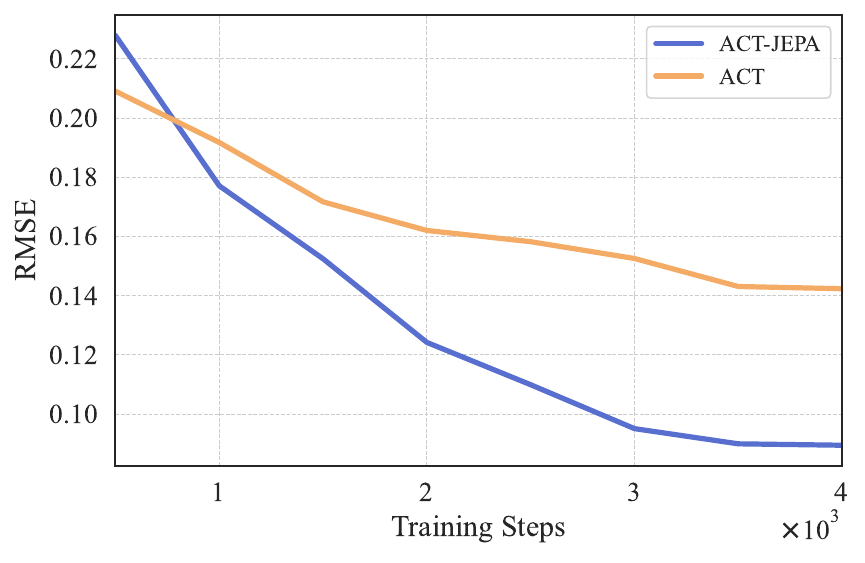}

        \vspace{2pt}
        {\footnotesize (a) ManiSkill RMSE $\downarrow$}
    \end{minipage}
    \hfill
    \begin{minipage}[b]{0.48\textwidth}
        \centering
        \includegraphics[width=\textwidth]{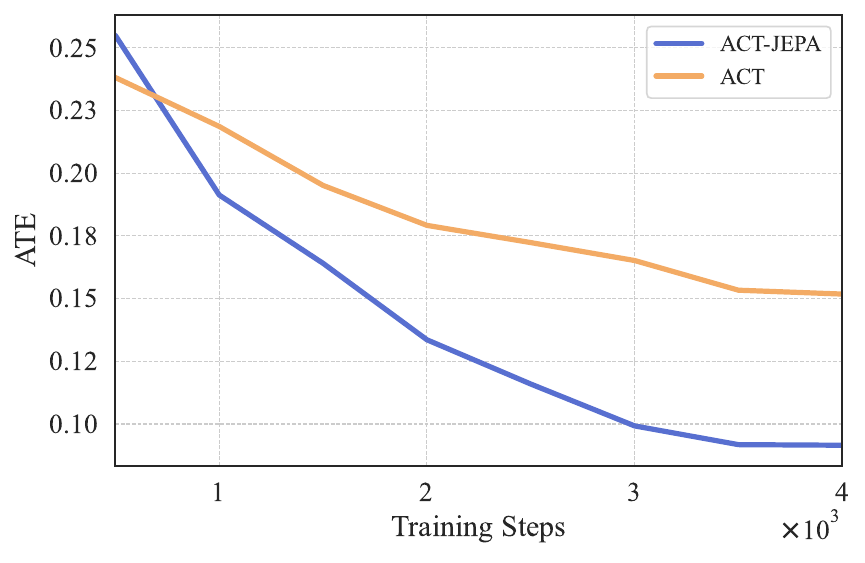}

        \vspace{2pt}
        {\footnotesize (b) ManiSkill ATE $\downarrow$}
    \end{minipage}
\caption{
	\textbf{World model representation quality during training.}
	RMSE (left) and ATE (right) for predicting future proprioceptive states throughout training in the ManiSkill environment.
	At fixed checkpoints, we freeze the context encoder and train a randomly initialized decoder to predict future states from the frozen representations, then evaluate its prediction error.
	Lower values indicate better predictive representations.
	ACT-JEPA reaches lower error earlier and maintains lower error overall, demonstrating more effective learning of environment dynamics.
}
    \label{fig:world_model_dynamics}
\end{figure*}

Results from the ManiSkill environment are shown in \autoref{fig:world_model_dynamics}; similar trends are observed in other environments.
Although ACT-JEPA may start with a higher error in both metrics, it quickly overtakes the baseline.
As training progresses, ACT-JEPA decreases prediction error faster, while ACT plateaus at higher error values.
This indicates that ACT-JEPA converges faster while achieving a lower error in both RMSE and ATE.
Predicting in latent space makes this approach efficient, aligning with recent JEPA works \cite{assran_self-supervised_2023,bardes_revisiting_2024}.

\begin{table*}[t]
	\centering
	\caption{\textbf{World model evaluation.}
	The quality of learned representations for world modeling is evaluated using a probing task.
	We freeze the encoder weights and train a decoder head on its representations to output sequences of future proprioceptive states.
	ACT-JEPA consistently outperforms the ACT baseline across benchmarks, achieving 29--40\% lower error, indicating that ACT-JEPA captures more information about environment dynamics.
	Lower values are better.
	}
	\label{tab:world_model_eval}
	\begin{tabular}{@{}lcccccc@{}}
		\toprule
		& \multicolumn{2}{c}{Push-T} & \multicolumn{2}{c}{ManiSkill} & \multicolumn{2}{c}{Meta-World} \\
		\cmidrule(lr){2-3}
		\cmidrule(lr){4-5}
		\cmidrule(lr){6-7}
		Method & RMSE $\downarrow$ & ATE $\downarrow$ & RMSE $\downarrow$ & ATE $\downarrow$ & RMSE $\downarrow$ & ATE $\downarrow$ \\ \midrule
		ACT & 0.1424 & 0.1518 & 0.0531 & 0.2063 & 0.0295 & 0.0529 \\
		ACT-JEPA & \textbf{0.0895} & \textbf{0.0915} & \textbf{0.0348} & \textbf{0.1354} & \textbf{0.0208} & \textbf{0.0375} \\ \bottomrule
	\end{tabular}
\end{table*}

\subsection{\RQtwo}
We evaluate ACT-JEPA's performance on decision-making tasks and benchmark it against the baseline policies.
Our primary metric is task success rate, defined as the percentage of successfully solved tasks in the environment.
Each policy is a transformer-based model trained from scratch on the same number of tokens per environment.
Policies are evaluated across diverse initial states per task to ensure robust evaluation.
For example, in the Meta-World environment, we evaluate 20 distinct initial states per task across 15 tasks ($20 \times 15 = 300$ total evaluations).
To isolate the effect of representation learning objectives and enable a fair comparison with ACT, ACT-JEPA uses the same action-decoder head as ACT. 
While more expressive decoders (e.g., diffusion-based) may improve performance in complex environments, our model naturally supports such alternatives, which we leave to future work.

We report the average task success rate results in \autoref{tab:eval-performance}.
ACT-JEPA outperforms all baselines across the selected environments.
Compared to the AR transformer policy, ACT-JEPA achieves an absolute improvement of +41\% in Push-T, +28\% in ManiSkill, and +53.7\% in Meta-World.
We attribute the AR policy's lower performance to compounding errors and non-Markovian behavior in the data, consistent with \cite{zhao_learning_2023}. 
In contrast, action chunking in ACT and ACT-JEPA mitigates these issues and improves performance.

Compared to the ACT policy, ACT-JEPA improves success rate by +7\% in Push-T, +10\% in ManiSkill, and +2\% in Meta-World.
The smaller improvement in Meta-World can be attributed to ACT already achieving near-ceiling performance (90\% success rate), leaving limited room for further gains.
We observe larger relative improvements of ACT-JEPA in environments where performance is not saturated, suggesting that learning a JEPA-based world model in addition to IL enhances policy robustness and generalization.
Overall, these results demonstrate that ACT-JEPA consistently improves decision-making performance across environments.

\begin{table*}[t]
	\centering
	\caption{\textbf{Policy performance.} Average task success rate (\%) across the evaluated benchmarks. Results show that ACT-JEPA consistently improves performance over imitation learning baselines across all environments.}
	\label{tab:eval-performance}
	\begin{tabular}{lccc}
		\toprule
		Method & Push-T (1 task) $\uparrow$ & ManiSkill (5 tasks) $\uparrow$ & Meta-World (15 tasks) $\uparrow$ \\ \midrule
		AR transformer & 0\% & 8\% & 38.3\% \\
		ACT & 34\% & 26\% & 90\% \\
		ACT-JEPA & \textbf{41\%} & \textbf{36\%} & \textbf{92\%} \\ \bottomrule
	\end{tabular}
\end{table*}

\subsection{\RQthree}\label{RQthree}
We isolate JEPA components and test if a learned world model captures policy-relevant features.
Specifically, we test if learning latent observation sequences transfers effectively to downstream tasks, such as predicting actions.
The assumption is that representations learned through one task (predicting observation sequences) can transfer to another (predicting action sequences), if there is an underlying connection between them.

To evaluate this, we design a two-stage training process: the model is first pretrained with SSL to learn latent observation sequences, after which the encoder is frozen and a newly initialized action decoder is trained with IL to predict action sequences.
In the pretraining stage, the goal is to learn useful representations that effectively transfer to downstream tasks, such as action prediction.
We utilize only JEPA components (context encoder, predictor, and target encoder) and discard the action decoder.
The model is pretrained to minimize the \(L_1\) loss between predicted and target latent observation sequences (\autoref{eq:observation_loss}).
In the second stage, we append a randomly initialized action decoder to the frozen encoder (as in ~\autoref{fig:overview}) and train it to reconstruct action sequences using the IL objective (~\autoref{eq:action_loss}).
To track representation quality over time, we periodically probe the encoder during pretraining. Every \(n\) steps, we freeze the encoder, train a new decoder from scratch, and record its action reconstruction loss and task success rate. The decoder is then discarded, and pretraining resumes with the original encoder.

\begin{figure}[ht]
    \centering
    \ifdim\columnwidth<\textwidth
        \includegraphics[width=0.99\linewidth]{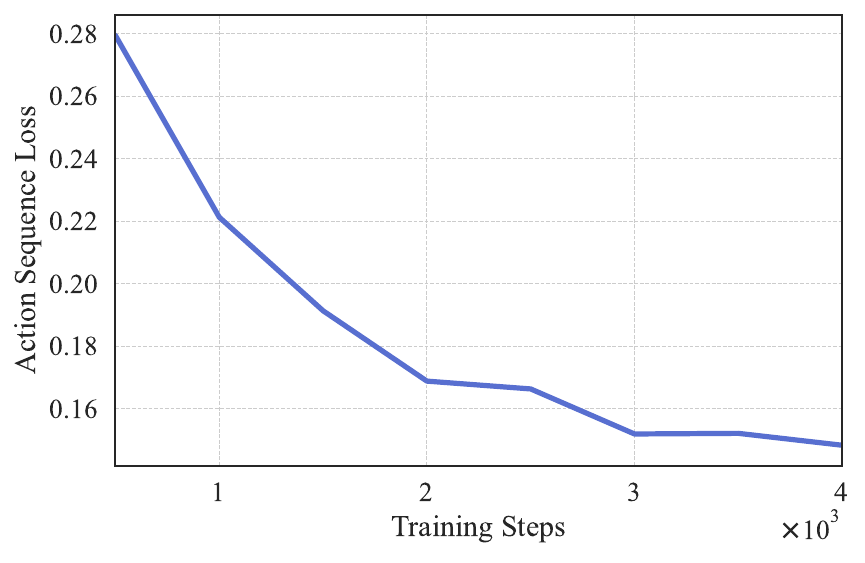}
    \else
        \includegraphics[width=0.6\linewidth]{transfer_action_loss_pusht}
    \fi
    \caption{\textbf{Action prediction probe during JEPA pretraining.}
    During SSL pretraining with JEPA, we periodically freeze the encoder, attach a newly initialized action decoder, and train the decoder to predict future action sequences.
    The figure shows results for the Push-T environment, while similar trends are observed in others.
    As pretraining progresses, the probing action decoder achieves lower reconstruction loss, showing that the learned latent dynamics become increasingly useful for downstream policy learning.
    }
    \label{fig: transfer to action prediction}
\end{figure}

As shown in \autoref{fig: transfer to action prediction}, the action reconstruction loss consistently decreases throughout training.
This indicates that the model gradually refines its internal representations during pretraining.
Although limited by dataset size, pretraining shows improvement in representation quality for downstream tasks.
This suggests that the learned world model could be utilized for different control tasks, including planning and alternative prediction objectives.
Importantly, the experiments confirm that learning a JEPA world model captures features useful for policy learning.

\subsection{\RQfour}
Combining SSL and SL is often implemented in a two-stage process: the model is first pretrained with SSL and then adapted to a downstream task with SL \cite{assran_self-supervised_2023,oquab_dinov2_2024,higuera_sparsh_nodate}.
This is also common in policy learning, where pretraining is followed by IL to learn actions \cite{urain_deep_2024,cui_dynamo_2024,black_0_nodate}.
In \autoref{RQthree} we demonstrated that such a two-stage approach is effective: pretraining captures representations useful for predicting both future states and actions.
However, it remains unclear whether optimizing both objectives separately is the most effective approach.
To investigate this, we compare the two-stage approach (\autoref{RQthree}) with an end-to-end approach that jointly optimizes SSL and IL objectives (\autoref{eq:final loss}).

As shown in \autoref{tab:joint_vs_two_stage}, joint optimization outperforms the two-stage approach on the reported benchmarks.
We attribute this performance gap to a representation misalignment in the two-stage approach.
During SSL pretraining, the model primarily focuses on modeling environment dynamics.
Although these representations are informative, they lack information required for fine-grained action generation.
Consequently, training only the action decoder on top of frozen representations cannot fully compensate for this mismatch, leading to worse performance than the end-to-end model.
This limitation is critical in robotics, where even small action errors can accumulate over time and cause task failure.

In contrast, joint optimization encourages the model to learn shared representations, allowing gradients from both objectives to shape the representation space.
By aligning both objectives throughout training, the representations become more coherent, robust, and generalizable, resulting in a more effective policy.
Joint optimization also provides a mutual regularization effect: the JEPA objective prevents overfitting to demonstration data by modeling environment dynamics, while the IL objective ensures the world model prioritizes control-relevant features for action generation.
It also simplifies the training pipeline by eliminating the need for multi-stage scheduling and managing fine-tuning phases.

These findings are particularly relevant for robotic systems developed for specific tasks under constrained data and computational resources.
In such settings, collecting demonstrations is costly and time-consuming, and deploying large pretrained models on robotic platforms may not be feasible.
From a practical perspective, replacing multi-stage scheduling with end-to-end training simplifies engineering pipelines.
How these findings extend to large-scale pretraining remains an open question. 
The two-stage approach may be advantageous when representations are reused across distinct downstream tasks, large-scale SSL data is available, or diverse trajectories (e.g., failure cases) can be leveraged to improve world model understanding.
However, when data and computational resources are limited, our results show that joint optimization is more effective than the conventional two-stage approach.

\begin{table*}[t]
	\centering
	\caption{\textbf{Comparison of two-stage and joint optimization approaches.} Average task success rate (\%) across benchmarks. In the two-stage approach, the model is first pretrained with JEPA, after which the encoder is frozen and a newly initialized action decoder is trained using IL. Joint optimization yields higher success rates in all three environments.}
	\label{tab:joint_vs_two_stage}
	\begin{tabular}{lccc}
		\toprule
		Method & Push-T (1 task) $\uparrow$ & ManiSkill (5 tasks) $\uparrow$ & Meta-World (15 tasks) $\uparrow$ \\ \midrule
		Two-stage approach & 27\% & 0\% & 23.3\% \\
		ACT-JEPA & \textbf{41\%} & \textbf{36\%} & \textbf{92\%} \\ \bottomrule
	\end{tabular}
\end{table*}

\subsection{Limitations}

Despite the significant theoretical implications of our findings, the scope of our empirical evaluation is limited by the benchmarks and available resources. 
Consistent with standard practice in the field, we evaluate the proposed architecture on common simulated benchmarks, which enable controlled and reproducible experimentation.
However, the evaluation remains limited in scale and data diversity compared to larger or real-world robotic settings.
We summarize the main limitations below.

\paragraph{Real-world environments.}
Without access to real-world robots, we could not deploy and test policies in real-world environments. Such environments present more complex challenges compared to the controlled conditions of simulations.

\paragraph{Target modality.}
We trained JEPA with proprioceptive states, leaving modalities such as touch, images, and depth unexplored.
While some modalities (e.g., tactile feedback) were unavailable, image data, although highly informative, are high-dimensional and require diverse and larger datasets to yield performance gains.

\paragraph{Dataset diversity.}
Simulated environments impose inherent limits on dataset diversity.
In real-world environments, lighting conditions and objects to manipulate can change significantly, producing a varied and diverse dataset, which is challenging in simulated environments.
For example, trajectories collected in Meta-World were recorded under uniform conditions: the same camera view, table, room, and lighting setup.
This homogeneity limits the dataset diversity needed to fully harness the advantages of JEPA.

\paragraph{Dataset size.}
Besides the lack of dataset diversity, our study is constrained by the dataset size.
Our datasets consist of hundreds of trajectories each, which is small compared to large-scale real-world datasets such as Bridge \cite{walke_bridgedata_2024} or Open X-Embodiment \cite{oneill_open_2024}.
As a self-supervised technique, JEPA is well suited to large and varied datasets.
We expect its most significant benefits to emerge as the number of trajectories, task diversity, and available modalities increase.
This aligns with trends in vision and language, where large-scale pretraining on broad datasets has led to remarkable success.
Scaling ACT-JEPA in this manner could enable it to serve as a foundation policy model.
How well the architecture transfers to larger, real-world datasets remains an open question that we plan to explore in future work.
\section{Conclusion}\label{conclusion}
In this work, we present ACT-JEPA, a novel end-to-end architecture that learns to generate actions using IL, while simultaneously learning a world model using JEPA.
By jointly optimizing these objectives, ACT-JEPA learns shared representations that improve both policy performance and world modeling.
Our experiments across multiple benchmarks demonstrate that ACT-JEPA consistently outperforms IL baselines in decision-making and world modeling tasks.
Additionally, learning a world model with JEPA yields representations that effectively transfer to downstream tasks.
While our experiments focus on IL, the learned world model opens opportunities for broader control applications, including planning and alternative prediction objectives.
In the future, we aim to scale the architecture with larger datasets and evaluate ACT-JEPA's performance in real-world environments.
Overall, these results highlight the advantage of integrating a JEPA-based world model with imitation learning, suggesting a promising direction for more effective and generalizable policies.
\section*{Acknowledgment}
This research has been supported by the Ministry of Science, Technological Development and Innovation (Contract No. 451-03-34/2026-03/200156) and the Faculty of Technical Sciences, University of Novi Sad through project "Scientific and Artistic Research Work of Researchers in Teaching and Associate Positions at the Faculty of Technical Sciences, University of Novi Sad 2026" (No. 01-3609/1).

\bibliographystyle{ieeetr}
\bibliography{references}

\end{document}